# Predicting Antibiotic Resistance Patterns Using Sentence-BERT: A Machine Learning Approach

Mahmoud Alwakeel, MD[1], Michael E. Yarrington, MD[1], Rebekah H. Wrenn, PharmD[1], Ethan Fang, Ph.D[2], Jian Pei, Ph.D[2], Anand Chowdhury, MD[1], An-Kwok Ian Wong, MD, Ph.D[1],

[1] Duke University School of Medicine, Durham, North Carolina, United States

[2] Department of Biostatistics & Bioinformatics, Duke University, Durham, North Carolina, United States

**Abstract:**

Antibiotic resistance poses a significant threat in in-patient settings with high mortality. Using MIMIC-III data, we generated Sentence-BERT embeddings from clinical notes and applied Neural Networks and XGBoost to predict antibiotic susceptibility. XGBoost achieved an average F1 score of 0.86, while Neural Networks scored 0.84. This study is among the first to use document embeddings for predicting antibiotic resistance, offering a novel pathway for improving antimicrobial stewardship.

**Introduction:**

Sepsis and septic shock are life threatening conditions, with mortality rates as high as 50-60%.[1] Delays in appropriate antibiotic administration lead to an 8% decrease in survival for every hour of delay, underscoring the need for prompt and precise treatment.[2] Despite this, only 50-60% of resistant cases receive appropriate therapy, and the overuse of broad-spectrum antibiotics remains high at 70%.[3,4] Inappropriate antibiotic use in these resistant cases doubles the risk of death.[4] Leveraging machine learning to analyze clinical documents and identify resistance patterns offers a novel approach not yet fully explored in the literature.

**Methods:**

We leveraged the MIMIC-III database to extract microbiological cultures and their antibiotic susceptibility patterns, alongside clinical notes recorded on or before the culture day. These notes were processed using Sentence-BERT (S-BERT) to generate embedding vectors that encapsulate the clinical context. Two predictive models, a Neural Network and XGBoost, were employed to classify antibiotic susceptibility as sensitive or resistant, including intermediate as resistant. The Neural Network model was built using a Multilayer Perceptron (MLP) with a single hidden layer of 100 neurons, ReLU activation, and the Adam optimizer. XGBoost was configured with 100 estimators, a learning rate of 0.3, a maximum depth of 6, and subsampling of 1. Model performance was evaluated using stratified 10-fold cross-validation to ensure balanced representation of outcome classes. Key metrics included Area Under the Curve (AUC) and F1 Score calculated for antibiotics, including Ceftriaxone, Zosyn, Meropenem, Ceftazidime, Amikacin, Tobramycin, and Vancomycin.

**Results**

Out of 29,440 available cultures, 28,074 were included in the study based on the presence of non-missing clinical documentation and the inclusion of cultures specific to bacterial organisms, excluding viral and other non-bacterial pathogens. The average age of the subjects was 62 years (SD 18). The cohort consisted of 55.4% males, and 70.8% of the entire cohort identified as White. The median number of clinical documentation entries per subject was 3 (IQR: 1-8). The cultures in the study were predominantly from respiratory sources (30.5%), followed by urinary (24.5%), and blood (17.0%). Swabs and screens accounted for 15.3%, while catheters and device-related cultures made up 4.4%. Fluid cultures represented 5.7%, and tissue and abscess samples comprised 3.9%. Other miscellaneous culture types contributed 2.3% of the total.

Cross-validation results indicated that XGBoost generally performed slightly better than Neural Networks, though both models exhibited similar metrics overall. XGBoost achieved an average AUC of 0.78 (SD 0.03) and an average F1 score of 0.86 (SD 0.03) across all tested antibiotics. In comparison, Neural Networks had an average AUC of 0.76 (SD 0.03) and an average F1 score of 0.84 (SD 0.04). XGBoost achieved the highest AUC of 0.81 and the highest F1 score of 0.91 for Meropenem. Neural Networks displayed a consistent AUC range from 0.71 (Amikacin) to 0.80 (Meropenem), with F1 scores ranging from 0.79 (Ceftazidime) to 0.91 (Meropenem).

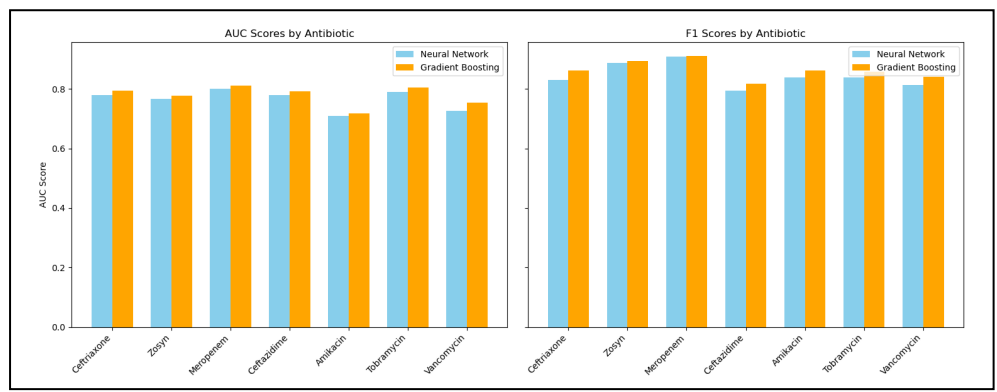

**Figure 1: Cross-Validation AUC and F1 Score Performance Across Various Antibiotics**

**Discussion:**
To our knowledge, this study is among the first to explore the use of machine learning tools, specifically document embeddings, to identify documentation phenotypes associated with antibiotic resistance patterns. Our study is strengthened by its large sample size and the diversity of cultures in the dataset. While the literature and guidelines have identified risk factors for resistant organisms, applying these stratifications has become increasingly challenging due to an aging population, advancements in medical care, frequent antibiotic exposure, prolonged hospitalizations, and the prevalence of patients in nursing facilities. These factors place patients at higher risk for resistant infections, often leading to the overuse of broad-spectrum antibiotics as a preventive measure.[3,5] Moreover, the high patient volume and reduced time per patient make it difficult to thoroughly review clinical documentation for risk stratification.[6] This highlights the value of utilizing machine learning and artificial intelligence to enhance patient phenotyping for appropriate antibiotic coverage. These tools could play a crucial role in ensuring effective treatment against resistant organisms during the critical first 48 hours before culture results are available. Additionally, they can assist in guiding antibiotic de-escalation or support decisions to continue broad-spectrum antibiotics when cultures are negative but the patient remains critically ill, as seen in approximately 50% of sepsis cases.[7] Limitations in our study included to be retrospective, single hospital system and confined one region which might not reflect the antibioticgram for other places. Also, our analysis depended on documentation which can vary significantly between providers or places, also other tubular data can carry important information does not exist in documentation. The next steps in our study include validating the model retrospectively and prospectively on Duke Health data. However, adoption of these tools at bedside may face challenges related to the lack of explainability; therefore, we plan to explore effective methods for providing clear reasoning and interpretations to support clinical decision-making.

**Conclusion:**
Utilizing S-BERT to generate embeddings from clinical notes demonstrates promising potential in enhancing the prediction of antibiotic resistance patterns. This approach underscores the value of document embeddings in clinical settings, presenting a novel pathway for improving decision-making in antimicrobial stewardship.